\documentclass[10pt]{article}
\usepackage[letterpaper]{geometry}
\usepackage{hicss}
\usepackage{times}
\usepackage[none]{hyphenat}
\usepackage{url}
\usepackage{latexsym}
\usepackage{minted}
\usepackage{indentfirst}
\usepackage{graphicx}
\graphicspath{{images/}}

\usepackage[style=apa, backend=biber]{biblatex}
\usepackage{enumitem}
\usepackage[most]{tcolorbox}
\usepackage{xcolor}
\usepackage{enumitem}

\title{Beyond RAG: A LLM-Based FAQ Matching Framework for Real-Time Decision Support in Contact Centers}

\author{
Garima Agrawal \quad Sashank Gummuluri \quad Cosimo Spera \\
Minerva CQ, Sunnyvale, California, USA \\
\texttt{\{garima, sashank, cosimo\}@minervacq.com}
}

\begin{document}
\maketitle
\begin{abstract}
In customer contact centers, human agents often face long average handling times (AHT) due to the need to manually interpret queries and search large knowledge bases (KBs). While retrieval-augmented generation (RAG) systems using large language models (LLMs) are increasingly adopted to support these tasks, they face limitations in real-time conversations—particularly with poorly formulated queries and repeated retrieval of frequently asked questions (FAQs). To address these issues, we propose a decision support framework that extends beyond RAG by combining real-time question identification with a dual-threaded FAQ matching and generation system. If the query matches a FAQ, the answer is retrieved instantly; otherwise a well-formed query is generated and routed to a RAG model. Deployed within Minerva CQ’s human-agent assist platform, our solution delivers sub-2-second responses for matched queries, significantly reduces unnecessary RAG calls, and lowers operational costs. We also introduce an automated, LLM-agentic pipeline for mining FAQs from historical transcripts, enabling continuous improvement of the FAQ knowledge base in the absence of manually curated QA pairs.
\end{abstract}

\subsubsection*{Keywords:}

Retrieval-Augmented Generation (RAG), RAG Optimization, Large Language Models (LLMs), FAQ Matching, Real-Time Decision Support, Contact Centers, Conversational AI, Cost Reduction

\section{Introduction}
\label{*Intro}

In today’s fast-paced customer contact centers, human agents must resolve inquiries efficiently while maintaining high service quality. One of the most critical metrics in this context is average handling time (AHT) \parencite{brantevica2018investigating}, which directly impacts operational costs and customer satisfaction. The longer it takes to resolve a query, the greater the risk of customer dissatisfaction.

Traditionally, human operators interpret customer queries manually and search large knowledge bases (KBs) for relevant content during live calls, often resulting in prolonged resolution times. To streamline this process, large language models (LLMs) have been integrated into customer service workflows, with retrieval-augmented generation (RAG) emerging as a popular approach \parencite{veturi2024rag, agrawal2024can}. RAG systems combine LLMs with retrieval mechanisms to surface relevant documents from KBs to support contact center agent interactions with customers. However, despite widespread adoption, RAG faces notable limitations in real-time environments \parencite{salemi2024evaluating, agrawal2024mindful}, including inaccurate query formulation and the repeated retrieval of frequently asked questions (FAQs).

Before detailing the challenges we encountered and our approach to addressing them, we briefly introduce Minerva CQ \parencite{minervacq}, an AI platform designed to assist human customer service agents in real time, where our solution is deployed.

Minerva CQ acts as an AI copilot, supporting service and sales agents by offering real-time assistance during live conversations. It embodies collaborative intelligence (CQ), where human and artificial intelligence work together to drive better outcomes than either could achieve independently. In this setup, AI supports human agents through adaptive workflows, dialogue suggestions, behavioral cues, and knowledge surfacing, all driven by real-time call transcripts.

We initially implemented a state-of-the-art RAG pipeline to surface relevant knowledge during customer calls. This pipeline leveraged Claude-3.5-Sonnet via Amazon Bedrock to query a curated corpus of company knowledge articles, including PDFs, JSON files, and plain text. For semantic search, Amazon OpenSearch Service was used for vector indexing, enriched with metadata such as document ID, source ID, page URL, and section markers. Articles originally available in formats like HTML, PHP, or other encodings were preprocessed and converted into structured JSON prior to ingestion and indexing.

Despite this architecture, human agents frequently reported that surfaced content lacked relevance to the ongoing conversation. Further analysis revealed that human operators were often inputting only basic keywords, which resulted in poor-quality queries and suboptimal retrieval. Additionally, we observed that the same questions were being asked across many calls, leading to repetitive and costly RAG operations.

These limitations underscored the need for a more intelligent, context-aware system capable of both improving query formulation and reducing redundant RAG calls. In this paper, we present an enhanced decision support system that addresses these challenges through two key innovations: (1) a real-time question identification model that suggests relevant questions based on conversational context, and (2) a dual-threaded pipeline that retrieves answers from an FAQ database when possible, or generates well-structured queries for RAG otherwise. If a question matches an existing FAQ, the corresponding answer is served instantly, bypassing the need for RAG inference. For non-FAQ cases, the system invokes RAG using a structured prompt generated from context.

To support this, we introduce a novel LLM-agentic pipeline for extracting FAQs from historical transcripts. This automated workflow ensures that organizations can maintain a dynamic and comprehensive FAQ database, even in the absence of predefined content.
Deployed within Minerva CQ’s AI-powered agent-assist platform for contact centers, our system demonstrates substantial efficiency gains, including faster response times, reduced average handling time (AHT), and significant cost savings through optimized RAG usage.

\noindent\textbf{Our contributions are as follows:}
\begin{enumerate}[noitemsep,topsep=0pt]
    \item \textbf{Enhanced Decision Support System}: We develop a real-time system for accurate question identification and answer generation to assist human agents during live conversations.
    \item \textbf{FAQ-Based RAG Optimization}: We introduce a dual-mode framework that retrieves answers from an FAQ database when possible and falls back on RAG otherwise, improving response latency, cost-efficiency, and answer quality.
    \item \textbf{Automated FAQ Mining Pipeline}: We design an LLM-agentic workflow to extract and curate FAQs from historical transcripts, enabling FAQ coverage without manual curation.
\end{enumerate}

\section{Solution Design}
\label{*Method}

\noindent \textbf{Problem Formulation:}  
Customer service agents spend valuable time manually searching large KB systems to address customer queries, increasing average handling time (AHT) and reducing efficiency \parencite{brantevica2018investigating}. The challenge is to automate this process by providing timely and accurate suggestions—either by retrieving answers from a predefined FAQ database or dynamically generating relevant questions based on the conversation context. This raises several key design questions:

\begin{itemize}[noitemsep,topsep=0pt] 
    \item How can we efficiently identify customer intent and surface relevant FAQs or KB articles in real time without disrupting the flow of conversation? 
    \item How can the system distinguish between queries that can be resolved via the FAQ database and those requiring RAG model retrieval? 
    \item How can we prevent redundant or already-answered questions while delivering contextually appropriate and timely responses? 
\end{itemize}

The objective is to reduce manual effort, lower AHT, and improve customer satisfaction by offering real-time, context-aware question suggestions and answers using a dual-threaded pipeline that integrates FAQ retrieval and RAG-based generation.

\subsection{FAQ Model}

We refer to our solution as the \textbf{FAQ Model}, designed to assist contact center human agents by offering real-time question suggestions and reducing the need to search KB articles manually. It maintains a curated database of FAQs, learned from historical transcripts, with corresponding answers that can either be pre-loaded or retrieved dynamically using a RAG model. During a live call, the model identifies the most relevant questions by analyzing recent conversational turns, while also ignoring those already addressed. It executes two parallel operations: \textbf{Match} and \textbf{Generate}.

The \textbf{Match} operation retrieves the top three most relevant questions from the FAQ database, based on the current context. Simultaneously, the \textbf{Generate} operation creates the top three questions dynamically using the same context window. These two threads present a unified list of six question suggestions, from which the human can selects the most appropriate. Depending on the source, the system retrieves the answer either directly from the FAQ database (for matched questions) or invokes the RAG model (for generated ones).

\subsubsection{Views in the FAQ Model}

The FAQ model supports four functional views tailored to different user roles and interactions:

\noindent\textbf{Agent-assist View:} During the call, human agents receive six suggested questions (three matched, three generated).
\begin{itemize}[noitemsep,topsep=0pt] 
    \item The model automatically filters out already-answered queries and surfaces only those requiring KB reference.
    \item Upon selecting a question, the corresponding answer is displayed in real time.
    \item Human agents can tag generated questions as new FAQs, which are then added to the FAQ database along with their answers.
\end{itemize}

\noindent \textbf{Supervisor View:} Supervisors have full control to view and manage the FAQ list and corresponding answers.
\begin{itemize}[noitemsep,topsep=0pt] 
    \item They can add, edit, or remove entries. If a question lacks an answer, supervisors can manually enter one or trigger a RAG-based response, which is then saved to the FAQ store.
\end{itemize}

\noindent \textbf{Model View:} The FAQ model can be invoked manually by agents or triggered automatically based on conversational heuristics.
\begin{itemize}[noitemsep,topsep=0pt]
    \item The \textbf{Match} and \textbf{Generate} threads run in parallel and return six suggestions.
    \item Based on the agent’s selection, the answer is retrieved from either the FAQ database or the RAG model.
\end{itemize}

\noindent \textbf{Data View:} All FAQs and their answers are stored in a persistent vector database for efficient retrieval.
\begin{itemize}[noitemsep,topsep=0pt]
    \item This database is editable by supervisors and allows runtime additions from agents.
\end{itemize}

\subsection{Key Advantages of the FAQ Model}

The FAQ Model provides multiple benefits across agent experience, performance, and cost-efficiency:

\begin{itemize}[noitemsep,topsep=0pt]
    \item \textbf{Human Agent Experience:} Automatically surfaces context-relevant question suggestions during calls, streamlining agent workflows. 
    \item \textbf{Reduced Time and Effort:} Removes the need for agents to formulate queries manually or browse KBs, providing immediate answers to frequently asked questions. 
    \item \textbf{Accuracy:} FAQ answers can be curated and validated by supervisors, while non-FAQ queries are handled through structured prompts sent to the RAG model—resulting in more accurate and context-aware responses. 
    \item \textbf{Cost Efficiency:} When a matched FAQ is available, the system bypasses the RAG model, reducing LLM API calls and overall inference costs \parencite{agrawal2024mindful}.
\end{itemize}

\section{Technical Implementation}
\label{*Implement}

This section outlines the technical implementation of our system, which consists of two primary components. First, we describe the offline pipeline for FAQ generation using historical transcripts. Second, we present the real-time model architecture for matching and generating questions during live calls.

\subsection{Generation of FAQs from Historical Transcripts}

When a predefined FAQ database is unavailable or insufficient, we initiate an automated pipeline to extract frequently asked questions from historical call transcripts using a set of five LLM-agents implemented with LangGraph. These AI agents process real-world turn-by-turn call transcripts collected from Minerva CQ’s deployed platform. A minimum of 500 call transcripts are typically sufficient to bootstrap FAQ generation in smaller deployments, while our current analysis uses approximately 30,000 such transcripts.

\begin{figure*}[t]
    \centering
    \includegraphics[width=0.60\textwidth]{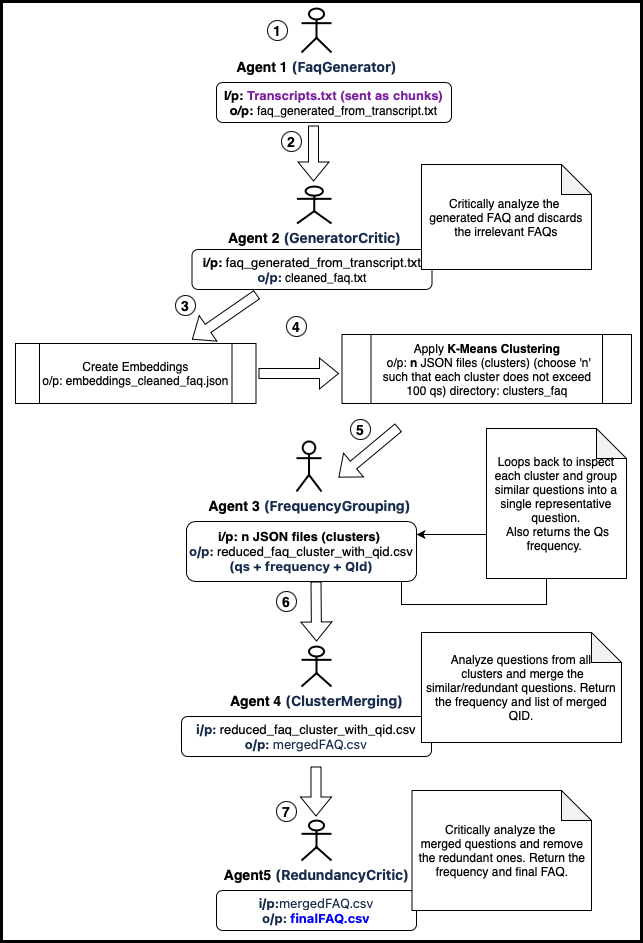}
    \caption{Agentic workflow for FAQ generation using five LLM agents to extract, cluster, and consolidate high-frequency customer questions from historical transcripts.}
    \label{fig:agentic_workflow}
\end{figure*}

The pipeline combines two key components: Amazon’s \texttt{titan-embed-text-v2} model (1024-dimensional embeddings) for clustering and Claude-3.5-Sonnet for semantic consolidation of representative questions. As illustrated in Figure~\ref{fig:agentic_workflow}, the workflow proceeds through the following steps:

\begin{itemize}
    \item \textit{Agent1} Question Generation (FaqGenerator): Segments transcripts into manageable chunks and extracts candidate customer questions.

    \item \textit{Agent2} Relevance Filtering (GeneratorCritic): Filters out irrelevant or non-informational utterances (e.g., greetings, verification prompts) to retain only actionable questions.

    \item \textit{Embeddings and Clustering:} Filtered questions are embedded using Amazon’s \texttt{titan-embed-text-v2} model and grouped with k-means clustering. We experimented with cluster sizes of $k=25, 40,$ and $60$, and found that $k=40$ produced the most coherent and representative clusters.
    \item \textit{Agent3} Frequency Grouping (Frequency Grouping): Implemented using Claude-3.5- Sonnet, this agent inspects each cluster and consolidates the questions into a single representative form while recording frequency data.

    \item \textit{Agent4} Cluster Merging (ClusterMerging): Representative questions across clusters are merged when semantically redundant.

    \item \textit{Agent5} Redundancy Critic (RedundancyCritic): Removes duplicates and performs quality filtering, producing a refined list of FAQs.
\end{itemize}

The output is a ranked list of high-frequency representative questions. We selected the top 100 and used the RAG pipeline to retrieve their corresponding answers. Both questions and answers are stored in a persistent vector database, enabling efficient retrieval during live calls.

\subsection{Real-Time Model Architecture and Execution}

Once the FAQ database is established, the system operates in real time to assist human agents using a dual-threaded pipeline composed of two tasks: \textbf{Match} and \textbf{Generate}.

The \textbf{Match} thread uses vector similarity to identify top-matching questions from the FAQ database, while the \textbf{Generate} thread dynamically produces candidate questions based on the most recent conversation turns. Together, they return six suggestions (three matched, three generated), enabling the human agent to select the most relevant one. If the selected question is an FAQ match, the corresponding answer is retrieved instantly; otherwise, the RAG pipeline is invoked to generate a response.

\subsubsection{Triggering Strategy}

One key challenge is determining when to invoke the model. We evaluated three strategies:
\begin{itemize}[noitemsep,topsep=0pt]
    \item \textbf{Manual Trigger:} The human agent explicitly clicks a button to request help.
    \item \textbf{Automatic Trigger:} The model monitors all turns and activates when a potential question is detected—this approach is resource-intensive.
    \item \textbf{Rolling Window Invocation:} We adopted a fixed-interval approach using a sliding window over conversation turns, balancing performance and efficiency.
\end{itemize}

\subsubsection{Implementation Approaches}

We evaluated three technical approaches to implement the Match and Generate threads in production:

\begin{itemize}[noitemsep,topsep=0pt]
    \item \textbf{Vector Search Only:} Matches conversation embeddings to the FAQ database. This method is fast (under 0.6 seconds) but requires a trained classifier to detect trigger points. Its performance degrades when user intent is distributed across multiple turns.
    
    \item \textbf{Large LLMs for Match + Generate:} Models like Claude-3.5-Sonnet~\parencite{claude2024} can perform both tasks in a unified pass. However, this approach introduced high latency (5.5+ seconds) and higher costs, making it unsuitable for real-time deployment.
    
    \item \textbf{Parallel Small LLMs:} We tested smaller models—Mistral-small 7B~\parencite{mistral2023}, LLaMA-8B~\parencite{llama2023}, Claude-3-Haiku~\parencite{claude2023}, and GPT-4o-mini~\parencite{openai2022}—to run the Match and Generate threads separately.
\end{itemize}

To select the optimal real-time model, we evaluated latency across multiple candidates under identical conditions. Claude-3-Haiku achieved the best balance, with response times consistently between 2.1–3.9 seconds and high-quality outputs. In contrast, Mistral-small and Mistral-7B exhibited moderate latencies (2.5–5.5s), while LLaMA-8B exceeded 5s in most cases. Larger models like Claude-3.5-Sonnet, though more capable, incurred latencies above 5.5s and were reserved for offline processing. These findings led us to deploy Claude-3-Haiku for real-time Match and Generate threads.

The final architecture uses a dual-thread Claude-3-Haiku configuration, running two lightweight models in parallel—each tuned for a specific subtask (matching or generating questions). To balance latency with conversational awareness, we invoke the Match and Generate threads at a rolling window of five conversation turns. Once the human agent selects a question from the suggested list, the system retrieves the corresponding answer from the FAQ store or, if unmatched, uses the RAG pipeline to fetch a response.

\subsection{Illustrative Example of Match and Generate Threads} \label{example}

Due to client confidentiality and privacy agreements, we are unable to present real call transcripts. However, we include the following fictional scenario to illustrate how the FAQ model's Match and Generate threads function in practice.

In this example, a customer contacts a service representative to inquire about a new Wi-Fi plan. At this point, the model is triggered (rolling window of 5 conversation turns), and both the Match and Generate threads are executed in parallel. A list of six relevant questions is suggested to the human agent—three retrieved from the FAQ database and three generated dynamically using the LLM. The agent selects the most appropriate question, and the system retrieves the corresponding answer.

\begin{tcolorbox}[
  colback=blue!5!white,
  colframe=blue!80!black,
  title=Conversation Snippet,
  fonttitle=\bfseries,
  before skip=4pt,
  after skip=4pt,
  boxrule=0.8pt,
  sharp corners
]
\textbf{Customer:} \textit{Hi, I’m moving next week and need to set up Wi-Fi at my new place.}\\
\textbf{Agent:} \textit{Sure, I can help with that. May I know the zip code of the new location?}\\
\textbf{Customer:} \textit{It’s 94086. I’d like something fast for work calls and streaming.}\\
\textbf{Agent:} \textit{Got it. I’ll check what we have available there.}\\
\textbf{Customer:} \textit{Do you have any offers or starter plans right now?}
\end{tcolorbox}

\begin{tcolorbox}[
  enhanced,
  colback=green!5!white,
  colframe=green!50!black,
  title=Match Thread Suggestions (from FAQ database),
  fonttitle=\bfseries,
  boxrule=0.8pt,
  sharp corners,
  width=\linewidth,
  left=2pt,
  right=2pt,
  top=2pt,
  bottom=2pt,
  before skip=6pt,
  after skip=6pt
]
\begin{itemize}[left=0pt, itemsep=2pt, label=--]
    \item What is included in the XYZ Starter Wi-Fi Plan?
    \item Are there any installation charges for new customers?
    \item What promotional offers are currently available for new connections?
\end{itemize}
\end{tcolorbox}

\begin{tcolorbox}[
  enhanced,
  colback=orange!5!white,
  colframe=brown!80!black,
  title=Generate Thread Suggestions (from LLM),
  fonttitle=\bfseries,
  boxrule=0.8pt,
  sharp corners,
  width=\linewidth,
  left=2pt,
  right=2pt,
  top=2pt,
  bottom=2pt,
  before skip=6pt,
  after skip=6pt
]
\begin{itemize}[left=0pt, itemsep=2pt, label=--]
    \item Can customers use their own router with the starter plan?
    \item Are there any data limits with the XYZ Starter Plan?
    \item Is a credit check required to activate a new plan?
\end{itemize}
\end{tcolorbox}

\noindent This example demonstrates how the dual-thread model reduces agent effort, surfaces high-quality contextual questions, and retrieves accurate responses—either through FAQ matching or, when needed, RAG generation.

\textbf{Note:} While we are unable to release source code due to enterprise deployment constraints, we are actively exploring the possibility of releasing a sanitized version to support research reproducibility.

\section{Results and Evaluation}
\label{sec:results}

\subsection{Key Observations}

The FAQ model was deployed in August 2024 within Minerva CQ's real-time agent assist platform. Since then, it has been invoked in over \textbf{80,000 customer calls} across multiple client deployments. 

Due to client data privacy restrictions, we cannot share actual conversation transcripts. However, our internal analysis and stakeholder feedback have revealed the following performance gains:

\begin{itemize}[noitemsep, topsep=2pt]
    \item \textbf{Improved RAG Response Quality:} Structured queries sent to the RAG model yielded more contextually relevant responses compared to earlier keyword-based queries. Domain experts and agents confirmed higher precision in retrieved answers.
    \item \textbf{Effective FAQ Generation:} The agentic workflow for mining FAQs from historical transcripts produced high-quality, reusable questions. Outputs were manually reviewed and validated by solution experts.
    \item \textbf{Relevance of Suggestions:} The Match and Generate modules consistently surfaced questions aligned with user intent. Iterative prompt refinement enhanced the quality and diversity of suggestions across customer intents.
\end{itemize}

\subsection{Evaluation Approach}

To assess model effectiveness over time, we propose tracking the following metrics:

\begin{itemize}[noitemsep, topsep=2pt]
    \item \textbf{Agent Interaction Ratio:} Percentage of conversations where agents selected a suggested question (matched or generated) versus typing a manual query.
    \item \textbf{Human Validation of FAQs:} Expert feedback on the clarity and relevance of FAQs generated from transcript data.
    \item \textbf{RAG Call Reduction:} Number of avoided RAG queries when a question was matched to the FAQ database, resulting in cost and latency savings.
\end{itemize}

\subsection{Preliminary Efficiency and Cost Analysis}
\label{sec:efficiency}

We analyzed logs from a recent 2-day window covering \textbf{$4,000$ enterprise calls.} Our FAQ-first architecture successfully intercepted \textbf{$7,391$ potential RAG queries} through real-time question matching.

\subsubsection{Cost Savings} 
Each RAG call typically uses \textbf{2,000 tokens} (1,000 input + 1,000 output). Based on Claude-3.5-Sonnet pricing (\$0.018 per call via AWS Bedrock) \cite{aws_bedrock_pricing}, we estimate:

\begin{itemize}[noitemsep,topsep=2pt,leftmargin=*]
  \item \textbf{7,391 RAG calls avoided} → \textbf{\$133 saved} in 2 days.
  \item Projected monthly savings: \textbf{\$1,996}.
  \item At company scale (10 clients): \textbf{\$19,960/month}, or over \textbf{\$240K/year} in LLM inference cost reduction.
\end{itemize}

\paragraph{Latency Reduction.}
Each avoided RAG invocation also cut agent wait time by an average of \textbf{6 seconds}. This translates to:

\begin{itemize}[noitemsep,topsep=2pt,leftmargin=*]
  \item \textbf{12.3 hours of cumulative agent latency saved} across the 2-day period.
  \item Projected monthly latency savings: \textbf{184.5 hours}.
\end{itemize}

\paragraph{Scaling Impact via FAQ Refresh.}
As customer questions evolve and KBs change, our system supports:
\begin{itemize}[noitemsep,topsep=2pt,leftmargin=*]
  \item \textbf{Promotion of new recurring queries} into the FAQ store.
  \item \textbf{Pruning of outdated entries} tied to obsolete KB facts.
  \item \textbf{Efficient batch regeneration} of FAQ answers when KB deltas are detected.
\end{itemize}

These compounding savings—both in cost and latency—make the framework highly scalable. Maintaining an up-to-date FAQ layer becomes essential to reduce future model invocations and support real-time response needs, ultimately ensuring long-term sustainability in LLM deployments.

\section{Related Work}
\label{*Related}

Recent research on real-time assistance systems has highlighted the transformative role of AI in reshaping the division of labor between humans and machines, particularly within the service industry. Studies such as \parencite{link2020use} examine both the opportunities and challenges introduced by AI-driven support platforms, which aim to reduce employee workload and enhance productivity by shifting certain decision-making responsibilities to advanced, real-time AI systems. One notable application area is customer support, where machine learning-powered chatbots and virtual assistants have shown effectiveness in automating routine queries and providing rapid, consistent responses \parencite{katragadda2023automating}.

As these technologies mature, the optimization of Natural Language Processing (NLP) techniques and Large Language Models (LLMs) has become central to enabling intelligent and hyper-personalized assistance. Integration of LLMs into service workflows supports a wide range of applications—including chatbots, sentiment analysis, question answering, and summarization—ultimately streamlining operations and improving the customer experience \parencite{kolasani2023optimizing}.

In particular, Retrieval-Augmented Generation (RAG) has emerged as a promising framework for improving the contextual accuracy of LLM outputs. Work such as \parencite{veturi2024rag} explores RAG-based question answering in customer service, while the RAGADA architecture \parencite{pitkaranta2024bridging} demonstrates how RAG can support more nuanced decision-making in enterprise environments by combining structured retrieval with generation.

However, several limitations persist in applying RAG to real-time customer interactions. For example, a recent evaluation of RAG-based chatbots \parencite{analytics2024retrieval} emphasizes the importance of retriever quality and prompt engineering in ensuring accurate system responses. The study finds that keyword generation strategies and retrieval configurations play a critical role in performance.

Despite these advances, traditional RAG systems face two notable challenges in practical deployment: (1) Human agents often struggle to formulate precise, context-rich queries during live conversations, which can degrade retrieval quality. (2) Frequently asked questions tend to be repeatedly routed to the RAG system by different agents, resulting in inefficiencies and redundant processing.

These limitations motivate the need for more context-aware, reusable solutions. The FAQ model proposed in this paper addresses both issues by enabling automatic identification of recurring questions and delivering instant responses through an FAQ retrieval mechanism—thereby reducing reliance on repeated RAG calls and improving real-time agent support.

\section{Conclusion}

This paper presents a novel decision support system that extends beyond traditional RAG approaches by optimizing their application in customer contact centers. The system addresses key challenges—such as poorly formulated queries and redundant retrieval of frequently asked questions (FAQs)—by integrating real-time question identification with RAG-based generation. This approach improves response accuracy, reduces average handling time (AHT), and enhances operational efficiency.

Deployed within Minerva CQ’s AI-powered human-agent assist platform, the system demonstrates strong potential to streamline agent workflows and reduce dependency on repeated RAG calls. In addition, we introduce an automated LLM-agentic workflow to identify FAQs from historical transcripts, enabling FAQ-based retrieval even when no predefined database is available.

Future work will focus on further extending the system’s capabilities beyond RAG, particularly in retaining multi-turn conversation history and handling more complex contextual queries. We also plan to evaluate its scalability across other domains, advancing research in real-time question identification and response generation at scale.



\printbibliography

\end{document}